\documentclass{article} 
\usepackage{iclr2025_conference,times}

\usepackage{hyperref}
\usepackage{url}
\usepackage{hyperref}
\usepackage{url}

\usepackage{amsmath}
\usepackage{amssymb}
\usepackage{mathtools}
\usepackage{amsthm}
\usepackage{enumitem}
\usepackage{xcolor}         
\usepackage{colortbl}
\usepackage[capitalize,noabbrev]{cleveref}

\usepackage{algorithm,algpseudocode}

\theoremstyle{plain}

\theoremstyle{definition}

\theoremstyle{remark}

\usepackage{cleveref}
\newcommand\myshade{85}
\colorlet{mylinkcolor}{violet}
\colorlet{mycitecolor}{violet}
\colorlet{myurlcolor}{violet}
\hypersetup{
  linkcolor  = mylinkcolor!\myshade!black,
  citecolor  = mycitecolor!\myshade!black,
  urlcolor   = myurlcolor!\myshade!black,
  colorlinks = true,
}

\usepackage[textsize=tiny]{todonotes}

\usepackage[compact]{titlesec}
\usepackage{sidecap}
\titlespacing*{\section}{0pt}{*0.15}{*0.15}
\titlespacing*{\subsection}{0pt}{*0.15}{*0.15}
\titlespacing*{\subsubsection}{0pt}{*0.1}{*0.1}

\usepackage{hyperref}       
\usepackage{url}            
\usepackage{booktabs}       
\usepackage{amsfonts}       
\usepackage{nicefrac}       
\usepackage{microtype}      
\usepackage{xspace}
\usepackage{paralist}

\usepackage{wrapfig}
\usepackage{amsmath}
\usepackage{amssymb}
\usepackage{amsthm}
\usepackage{url}
\usepackage{paralist}
\usepackage{latexsym}
\usepackage{arydshln}
\usepackage{tabularx}
\usepackage{verbatim}
\usepackage{CJK}
\usepackage{flushend}
\usepackage{multicol}
\usepackage{multirow,makecell}
\usepackage{color}
\usepackage{colortbl,array}
\usepackage{silence}
\usepackage{arydshln} 
\usepackage{xspace}
\usepackage{soul}
\usepackage{pifont}

\usepackage{arydshln}

\usepackage{amsmath,amsfonts,bm}

\def\Figref#1{{Fig.}~\ref{#1}}

\def\Secref#1{{\S}\ref{#1}}

\def\eqref#1{equation~\ref{#1}}
\def\Eqref#1{{Eq.}~(\ref{#1})}

\def\Tabref#1{{Tab.}~\ref{#1}}

\def\1{\bm{1}}

\def\vs{{\bm{s}}}

\DeclareMathAlphabet{\mathsfit}{\encodingdefault}{\sfdefault}{m}{sl}
\SetMathAlphabet{\mathsfit}{bold}{\encodingdefault}{\sfdefault}{bx}{n}

\newcommand{\R}{\mathbb{R}}

\usepackage{tikz}

\makeatletter
\DeclareRobustCommand\onedot{\futurelet\@let@token\@onedot}
\def\@onedot{\ifx\@let@token.\else.\null\fi\xspace}

\def\eg{\textit{e.g}\onedot} 
\def\ie{\textit{i.e}\onedot} 
 
 \def\vs{\textit{vs}\onedot}

\def\wrt{\textit{w.r.t}\onedot} 

\makeatother

\newcommand{\hidethis}[1]{}

\definecolor{bblue}{HTML}{4F81BD}
\definecolor{oorange}{HTML}{F4C842}
\definecolor{rred}{HTML}{C0504D}
\definecolor{ggreen}{HTML}{9BBB59}
\definecolor{ppurple}{HTML}{9F4C7C}
\definecolor{darkgreen}{HTML}{228B22}
\definecolor{cred}{HTML}{D81B60}
\definecolor{cblue}{HTML}{1E88E5}
\definecolor{cyellow}{HTML}{FFC107}
\definecolor{nred}{HTML}{e41a1c}
\definecolor{nblue}{HTML}{377eb8}
\definecolor{ngreen}{HTML}{4daf4a}
\definecolor{lblue}{HTML}{6C8EBF}

\newcommand{\cmark}{\ding{51}}%
\renewcommand{\checkmark}{\cmark}

\newlength\savewidth

\newcolumntype{x}[1]{>{\centering\arraybackslash}p{#1pt}}
\newcolumntype{y}[1]{>{\raggedright\arraybackslash}p{#1pt}}
\newcolumntype{z}[1]{>{\raggedleft\arraybackslash}p{#1pt}}

\newcommand{\app}{\raise.17ex\hbox{$\scriptstyle\sim$}}

\definecolor{deemph}{gray}{0.6}

\definecolor{baselinecolor}{gray}{.9}

\definecolor{emerald}{rgb}{0.31, 0.78, 0.47}
\definecolor{Gray}{gray}{0.9}
\definecolor{Highlight}{rgb}{0.99,0.97,0.93}
\usepackage[first=0,last=9]{lcg}
\newcommand{\chl}{\cellcolor{Highlight}}

\renewcommand{\paragraph}[1]{\noindent\textbf{#1}}
\renewcommand{\bm}[1]{\mathbf{#1}}

\newcommand{\method}{\textsc{DPLM-2}\xspace}

\newcommand{\xt}[1]{\bm{x}^{(#1)}}
\newcommand{\metric}[1]{\texttt{#1}}

\title{DPLM-2: A Multimodal Diffusion Protein Language Model}

\author{Xinyou Wang\thanks{This work was done during Xinyou's internship at ByteDance Research.}~~\!$^{\diamondsuit\!\heartsuit}$\!
Zaixiang Zheng\thanks{Project Lead.}~~\!$^\heartsuit$ 
Fei Ye$^\heartsuit$ 
Dongyu Xue$^\heartsuit$
Shujian Huang$^\diamondsuit$
Quanquan Gu\thanks{Corresponding Author.}~~\!$^\heartsuit$ \\
$^\diamondsuit$Dept. of Computer Science, Nanjing University\\
$^\heartsuit$ByteDance Research \\
{$\mathtt{wangxinyou@smail.nju.edu.cn,~~\{zhengzaixiang,quanquan.gu\}@bytedance.com}$}\\
{Project Page: \url{https://bytedance.github.io/dplm/dplm-2}}\\
}

\iclrfinalcopy 
\begin{document}

\maketitle

\vspace{-3mm}
\begin{abstract}

Proteins are essential macromolecules defined by their amino acid sequences, which determine their three-dimensional structures and, consequently, their functions in all living organisms. Therefore, generative protein modeling necessitates a multimodal approach to simultaneously model, understand, and generate both sequences and structures. However, existing methods typically use separate models for each modality, limiting their ability to capture the intricate relationships between sequence and structure. This results in suboptimal performance in tasks that requires joint understanding and generation of both modalities.
In this paper, we introduce DPLM-2, a multimodal protein foundation model that extends discrete diffusion protein language model (DPLM) to accommodate both sequences and structures.
To enable structural learning with the language model, 3D coordinates are converted to discrete tokens using a lookup-free quantization-based tokenizer.
By training on both experimental and high-quality synthetic structures, DPLM-2 learns the joint distribution of sequence and structure, as well as their marginals and conditionals.
We also implement an efficient warm-up strategy to exploit the connection between large-scale evolutionary data and structural inductive biases from pre-trained sequence-based protein language models.
Empirical evaluation shows that DPLM-2 can simultaneously generate highly compatible amino acid sequences and their corresponding 3D structures eliminating the need for a two-stage generation approach.
Moreover, DPLM-2 demonstrates competitive performance in various conditional generation tasks, including folding, inverse folding, and scaffolding with multimodal motif inputs, as well as providing structure-aware representations for predictive tasks.



\end{abstract}


\section{Introduction}
\label{sec:intro}

Proteins are macromolecules that execute crucial roles in every living organism.
They are characterized by their amino acid sequences and three-dimensional structure, where the sequence determines the structure, which in turn governs the
protein’s function.
Generative modeling for proteins has made significant strides in recent years.
Among them, diffusion models~\citep{ho2020ddpm, song2020sde} exhibit great success in protein structure-based generative modeling~\citep{watson2023RFdiffusion, yim2023framediff}. 
Meanwhile, large-scale protein language models~\citep{rives2019esm,lin2022esmfold}, trained on evolutionary-scale sequence database, have become one of the most important cornerstones in sequence-based foundation models for protein sequence representation learning and generation. 
Remarkably, DPLM~\citep{wang2024diffusion}, a discrete diffusion~\citep{austin2021structured} based protein language models, has exhibited the state-of-the-art performance in both sequence generation and understanding, addressing a wide range of sequence-oriented applications.

Many protein engineering applications, \eg, motif-scaffolding~\citep{watson2023RFdiffusion,yim2024improved} and antibody design~\citep{jin2021iterative,kong2022conditional,zhou2024abdpo}, require jointly determine both structure and sequence. 
However, the aforementioned approaches mostly employ generative models for one modality (either sequence or structure) and resort to separate models~\citep{jumper2021AF2,dauparas2022proteinmpnn} for the other.
This highlights the pressing need for multimodal protein generative models that can integrate both sequence and structure, enabling a more comprehensive understanding of protein behaviors and functions. 
This, therefore, raises the following question:
\begin{center}
    \vspace{-2pt}
    {\it Can we build a multimodal protein foundation model to simultaneously \\ model, understand, and generate both sequences and structures?}
    \vspace{-2pt}
\end{center}

To pursue this goal, Multiflow~\citep{campbell2024generative} is a recent effort for structure-sequence co-generation that incorporates sequences into structure-based generative models using multimodal flow matching. 
Despite its impressive structure generation capability, Multiflow exhibits suboptimal performance in co-generating structurally-compatible sequences and consequently resorts to instance-level knowledge distillation from ProteinMPNN~\citep{dauparas2022proteinmpnn}. 
Furthermore, it completely falls short in protein folding for given sequences, showing Mulitflow's inadequacy in sequence understanding.
We argue that this bottleneck arises from the absence (co-)evolutionary inductive bias derived from massive pre-training from sequence database, as prior studies have demonstrated that the evolutionarily-informed representations learned by pre-trained protein language models implicitly capture structural information enables direct structure prediction~\citep{lin2022esmfold}.
As a consequence, the limitation in sequence understanding and generation renders Multiflow inadequate as a multimodal protein generative foundation.

\begin{figure*}[t]
    \vspace{-5.5mm}
    \centering
    \includegraphics[width=\linewidth]{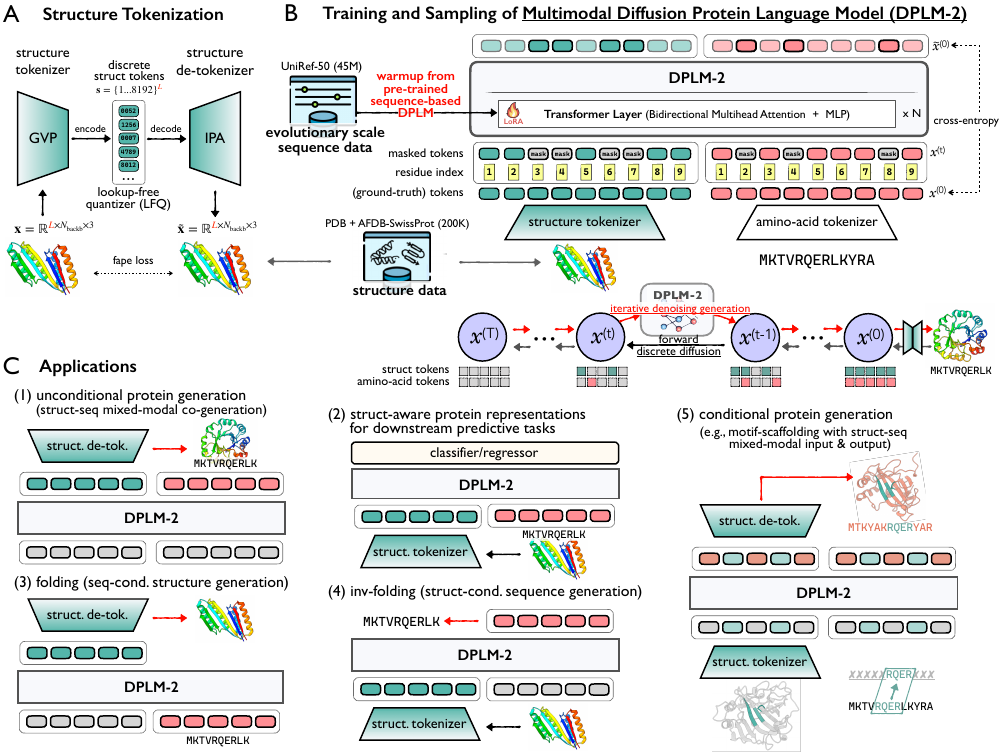}
    \vspace{-3mm}
    \caption{{\sl Overall illustration of \method.}
    \textbf{(A)} Structure tokenization consists of a GVP-based encoder to yield invariant backbone geometric features, a lookup-free quantizer (LFQ) to discretize encoded structural features into structure tokens within a codebook, and an IPA-based decoder as de-tokenizer to convert structure tokens back to backbone atomic coordinates. 
    \textbf{(B)} Multimodal learning and generation of protein structure and sequence with DPLM-2.
    \textbf{(C)} Various applications of DPLM-2 as a protein foundation model: (1) unconditional protein sequence-structure mixed-modal co-generation; (2) protein sequence-structure joint representation for predictive tasks; (3) structure prediction; (4) fixed-backbone sequence generation; (5) conditional protein generation with structure-sequence mixed-modal input and output.
    }
    \label{fig:main}
    \vspace{1mm}
\end{figure*}

Inspired by the connection between evolutionary knowledge and spatial interactions, we deem that sequence-based generative language models like DPLM, with their strong sequence generation and predictive abilities, hold great promise as a foundation for multimodal learning for proteins. 
Despite its exciting potential, this approach presents two key challenges: (1) language models cannot directly handle continuous data like structure; and (2) language models heavily necessitate sufficient scale of data and compute resources while structure data is much smaller compared to sequence databases.

In this paper, we address the aforementioned questions by introducing DPLM-2, a multimodal protein foundation model that advances the state-of-the-art discrete diffusion-based protein language model (\ie, DPLM) to accommodate both sequences and structures.
By training on both experimental and high-quality synthetic structures, DPLM-2 learns the joint distribution of sequence and structure, as well as their marginals and conditionals.
We present several key recipes to facilitate multimodal learning in DPLM-2: 
(1) the core difficulty lies in enabling the language model to learn structural information, which is challenging and remains elusive, for which we develop a lookup-free quantization~\citep[LFQ,][]{yu2023language} structure tokenizer to convert 3D coordinates to discrete tokens and vice versa (\Figref{fig:main}A, \S\ref{sec:method-tokenizer});
(2) we implement an efficient warm-up strategy to exploit the connection between large-scale evolutionary data and structural inductive biases from pre-trained sequence-based DPLM (\Figref{fig:main}B, \S\ref{sec:method-train});
and (3) we also address the exposure bias problem in discrete diffusion for sequence learning~\citep{ranzato2015sequence,bengio2015scheduled} by a self-mixup training strategy that leads to enhanced generation quality and diversity.

We highlight our main contributions and findings as follows:
\begin{compactitem}

    \item[(i)] 
    We present \method, a multimodal protein generative language model that aims to simultaneously model, understand and generate protein structure and sequence.
    We show that it can be fairly efficient and effective to obtain a mulitmodal protein model with moderate amount of high-quality data, a decent structure tokenizer and publicly-accessible sequence-only pre-trained language models.
    
    \item[(ii)] 
    As a mulitmodal generative model, \method enables unconditional co-generation of designable and diverse proteins that guarantees consistency between structure and sequence~(\Figref{fig:main}C(1)). 
    Our empirical evaluation shows that \method attains competitive co-generation performance compared to structure-based generative approaches, while the  proteins generated by \method have a better alignment with the characteristics of natural proteins in secondary structure statistics (\S\ref{sec:exp-uncond}). 
    
    \item[(iii)] In addition, \method supports various conditional generation tasks by its multimodal nature, ranging from (sequence-conditioned) folding~(\Figref{fig:main}C(3), \S\ref{sec:exp-folding}), (structure-conditioned) inverse-folding~(\Figref{fig:main}C(4), \S\ref{sec:exp-invfold}), to more successful motif-scaffolding given multimodal motif conditioning~(\Figref{fig:main}C(5), \S\ref{sec:exp-motif}). 
    
    \item[(iv)] Last but not least, we demonstrate that the structure-aware protein representation learned by \method brings additional benefit for a range of protein predictive tasks (\Figref{fig:main}C(2), \S\ref{sec:exp-repr}).

\end{compactitem}

\paragraph{Concurrent work.} 
During the development of \method, we became aware of the recently proposed multimodal generative protein language model, ESM3~\citep{hayes2024esm3}, which also jointly models tokenized structure and sequence using a generative masked language model. While both models aim for similar goals, \method differs from ESM3 in several key aspects:
\textbf{(1)} {\it Multimodal protein generation:} \method treats structure and sequence modalities equally by design and emphasizes the simultaneous co-generation of compatible protein sequence and structure, whereas ESM3 is a sequence-first model (other modalities are subject to dropout during training) and generates in cascaded modality-by-modality manner.
\textbf{(2)} {\it Data and compute efficiency:}
ESM3 seeks to perform mulimodal pre-training from scratch using a huge amount of synthetic data, with modal size ranging from 1.4B to 98B.
With strict license and absence of training infrastructure, this prohibits community from replicating for customized purposes.
In contrast, \method leverages much smaller datasets~(PDB + SwissProt) and builds on open-source, pre-trained sequence-based DPLM (150M/650M/3B), which leverages DPLM's learned evolutionary knowledge and inherits strong sequence understanding and generation capabilities. 
We are also committed to open-source our models, training and inference code to democratize multimodal generative protein LMs to benefit the community. 
Overall, we believe \method provides unique contributions to the community.

\section{Preliminaries}
\label{sec: preliminary}
\subsection{Generative Modeling for Protein}

\begin{wraptable}[7]{r}{0.28\textwidth}
\centering
\footnotesize
\vspace{-9mm}
\setlength{\tabcolsep}{4.5pt}
\caption{{\sl Generative tasks \wrt structure \& sequence.}
}
\label{tab:task}
\vspace{1.5pt}
\resizebox{\linewidth}{!}{%
\begin{tabular}{ll}
\toprule
 task &  objective  \\
\midrule
folding &  $p_\theta( \bm{x} | \bm{s})$  \\
inv-folding &  $p_\theta(\bm{s} | \bm{x})$  \\
seq. gen. &  $p_\theta(\bm{s})$  \\
struct. gen. &  $p_\theta(\bm{x})$  \\
\midrule
seq-struct co-gen. &  $p_\theta(\bm{s}, \bm{x})$  \\
\bottomrule
\end{tabular}
}
\end{wraptable}
The aim of generative protein modeling is to estimate the underlying distribution $\mathrm{prot} \sim q(\mathrm{prot})$ of the protein data of our interest by learning a probabilistic model $p_\theta(\mathrm{prot})$.
Here $\mathcal{\mathrm{prot}} = (r_1, r_2, \dots, r_L) $ denotes a protein with $L$ residues, where each residue $r_i = (s_i, x_i)$ is represented by two major modalities, \ie, $s_i \in \{0,1\}^{ |\mathcal{S}|}$ is a categorical variable for its amino acid type in $\mathcal{S} = \{1,..., 20\}$, and $x_i \in \R^{N_\text{atoms} \times 3}$ is the real-value Cartesian coordinates of its residue atoms (we only consider backbone atoms herein, \ie, $[\text{N}, \text{C}_{\alpha}, \text{C}, \text{O}]$ with $N_\text{atoms}=4$).
Namely, 
\begin{align}
    p_\theta(\mathrm{prot}) = p_\theta(s_1, s_2,\dots, s_L,~ x_1, x_2, \dots, x_L) = p_\theta(\bm{s}, \bm{x}) \nonumber
\end{align}
As a result, most of protein tasks can be viewed as specifying their input conditioning and output between these two modalities (\Tabref{tab:task}), including 
(1) sequence-conditioned structure prediction~\citep[folding,][]{jumper2021AF2,lin2022esmfold,huguet2024foldflow2}, 
(2) structure-conditioned sequence generation~\citep[inverse folding or fixed-backbone design,][]{dauparas2022proteinmpnn,hsu2022esmif,zheng2023structure},
(3) sequence learning or generation ~\citep{rives2019esm,nijkamp2022progen2,alamdari2023protein,wang2024diffusion},
(4) structure generation~\citep{yim2023framediff,watson2023RFdiffusion,ingraham2023chroma},
and (5) sequence-structure co-generation~\citep{jin2021iterative,shi2022protein,campbell2024generative}.
These further enable various conditional applications by allowing single or mixed-modal conditioning for partial generation, \eg, motif-scaffolding and antibody design.

\subsection{Diffusion Protein Language Model (DPLM)}
Language models (LMs), typically parameterized by Transformers~\citep{vaswani2017attention} have become the \emph{de facto} choice dominating different domains with scalable and performing expressiveness~\citep{openai2023gpt4}.
Among them, protein LMs have been serving as one of the AI foundation for protein sequence learning~\citep{rives2019esm,lin2022esmfold} and generation~\citep{nijkamp2022progen2,alamdari2023protein}. 

Diffusion protein language model~\citep[DPLM,][]{wang2024diffusion}, in particular, shows excelling performance in both generation and representation learning of protein sequences.
DPLM is grounded in \textit{absorbing} discrete diffusion framework~\citep{austin2021structured, zheng2023reparameterized}, which is characterized by a forward and backward Markov process.
Let $\texttt{Cat}(\bm{x};\bm{p})$ be a categorical distribution on protein sequence $\bm{y}$ parameterized by a vector $\bm{p}$ on $(|\mathcal{V}|-1)$-dimensional probability simplex.
The forward process of discrete diffusion defines a Markov process governed by the transition kernel
$q(\xt{t}|\xt{t-1})=\texttt{Cat}\big(\xt{t}; \beta_t\xt{t-1} + (1-\beta_t)\bm{q}_{\text{noise}}\big)$
that gradually perturb the data $\xt{0}\sim q(\xt{0})$ into a stationary distribution $\xt{T} \sim \bm{q}_{\text{noise}}$. 
For absorbing diffusion, $\bm{q}_{\text{noise}}$ is the point mass with all of the probability on the absorbing (mask) state.
The learned \textit{backward} process $p_{\bm{\theta}}(\xt{t-1}|\xt{t})$ reversely denoises the $\xt{T}$ towards the data distribution $\xt{0}$, which is typically optimized by the variational bound of the log-likelihood~\citep{ho2020ddpm}:
\begin{align}
      & \mathbb{E}_{q(\xt{0})}\big[\log p_\theta(\xt{0})\big]  \geq \mathbb{E}_{q(\xt{0:T})} \bigg[\log \frac{p_{\theta}(\xt{0:T})}{q(\xt{1:T}|\xt{0})}\bigg] \nonumber\\[-5pt]
      & = \mathbb{E}_{q(\xt{0})}\Big[\log p_{\theta} (\xt{0} | \xt{1}) 
      + \textstyle{\sum_{t=2}^{T}} \underbrace{-\text{KL}\big[q(\xt{t-1}|\xt{t}, \xt{0})\|p_{{\theta}}(\xt{t-1}|\xt{t})\big]\Big]}_{\mathcal{J}_t} + \text{const.}, \nonumber
\end{align}
where $\mathcal{J}_t$ is the learning objective. 
The learning objective of discrete diffusion can be further simplified into reweighted cross-entropies~\citep{zheng2023reparameterized}, resembling masked language modeling at arbitrary noise levels:
\begin{align}
\mathcal{J}_t & = \mathbb{E}_{q(\xt{0})}-\text{KL}\big[q(\xt{t-1}|\xt{t}, \xt{0})\|p_{{\theta}}(\xt{t-1}|\xt{t})\big] \nonumber \\[-1pt]
& = \mathbb{E}_{q(\xt{0})} \Big[\lambda^{(t)}  \textstyle{\sum_{1 \leq i \leq L}} b_i(t) \cdot \log p_{\theta}(x^{(0)}_i|\xt{t})\Big], 
\label{eq:reparam_obj}
\end{align}
where $\lambda^{(t)}$ is a weighting coefficient induced from the specific noising schedule. 
For inference, DPLM is able to generate amino acid sequences by the reverse iterative denoising process of discrete diffusion~\citep{hoogeboom2021argmax,austin2021structured} from the following distribution,
\begin{align}
    p_\theta(\xt{t-1} | \xt{t}) = \textstyle\sum_{\Tilde{\bm{x}}^{(0)}} q(\xt{t-1} |\xt{t}, \Tilde{\bm{x}}^{(0)} p_\theta(\Tilde{\bm{x}}^{(0)}| \xt{t}). \nonumber
\end{align}
Specifically, at time $t$, it first generates $\Tilde{\bm{x}}^{(0)}$ from $p_\theta(\cdot| \xt{t})$, then a less noisy $\xt{t-1}$ is sampled by $q(\cdot |\xt{t},\xt{0} = \Tilde{\bm{x}}^{(0)})$.
Within absorbing diffusion, the generation process can be viewed as an iterative \textit{mask-predict} approach.
For sequence representation for predictive tasks, it can be obtained by simply letting DPLM take the sequence as input.











\section{DPLM-2: A Multimodal Diffusion Protein Language Model}

\subsection{Overview}
\Figref{fig:main} illustrates \method's overall architecture.
\method is built on the state-of-the-art sequence-based generative protein LM, \ie, DPLM~\citep{wang2024diffusion}, using a discrete diffusion probabilistic framework to concurrently model both protein sequences and their corresponding structures.
To facilitate structure learning in language models, we introduce a token-based representation for protein structure via a tokenizer that converts \(\mathbf{x} \in \mathbb{R}^{L \times N_\text{backb} \times 3}\), the 3D coordinates of the protein backbone into a discrete structure token sequence, denoted as \(\bm{z} = (z_1, z_2, \dots, z_L) \in \{0,1\}^{L \times |\mathcal{Z}|}\), where each token \(z_i\) represents a local structural element of the \(i\)-th residue.
Given tokenized structure, \method processes mulitmodal input by concatenating the structure token sequence \(\bm{z}\) with the corresponding amino acid sequence \(\bm{s}\) for the same protein.
Notably, there exists a position-by-position correspondence between \(\bm{z}\) and \(\bm{s}\), where \(z_i\) and \(s_i\) refer to the two modalities of the $i$-th residue, respectively. 
To reinforce this correspondence, we assign identical position encodings to both \(z_i\) and \(s_i\), thereby ensuring that structural and sequence information is aligned at the residue level.


To train \method, we leverage a high-quality dataset comprising 20K clustered experimental structures from the Protein Data Bank (PDB)~\citep{berman2000protein} and 200K predicted structures from the AFDB SwissProt split~\citep{varadi2022alphafold}, with length $< 512$.
During training, \method is tasked with denoising the input sequence across a spectrum of noise levels, ranging from fully noisy to completely clean. 
The multimodal training objective of \method is derived from \Eqref{eq:reparam_obj} as,
\begin{align}
\mathcal{J}_{t} & = \mathbb{E}_{q(\bm{x}^{(0)},\bm{s}^{(0)}),\bm{z}^{(0)}\leftarrow \textit{tokenize}(\bm{x}^{(0)})} \Big[\lambda^{(t)}  \textstyle{\sum_{1 \leq i \leq L}} b_i(t) \cdot \log p_{\theta}(z^{(0)}_i,s^{(0)}_{i}|\bm{z}^{(t)},\bm{s}^{(t)})\Big], \nonumber 
\end{align}





where $\log p_{\theta}(z_i,s_i|\cdot) = \log p_{\theta}(z_i|\cdot) + \log p_{\theta} (s_i|\cdot)$ by assuming conditional independence. By learning \(p_{\theta}(\bm{z}^{(t-1)}, \bm{s}^{(t-1)} | \bm{z}^{(t)}, \bm{s}^{(t)})\), the model enables the simultaneous generation of highly correlated protein structures and sequences. This eliminates the need for a cascaded generation paradigm, allowing us to derive both the protein's structure and sequence in a single step.

To further enhance \method's ability to differentiate between structure and sequence, noising level for each modality is subjected to distinct scheduler, denoted as \(t_{\mathbf{z}}\) and \(t_{\mathbf{s}}\), respectively. 
This facilitates a more comprehensive understanding of the relationships between protein sequences and their corresponding structures.
This design also allows us to explore arbitrary combinations of \((t_{\mathbf{z}}, t_{\mathbf{s}})\), thus providing flexible sampling options, including sampling from the marginals of each modality and conditionals between them for various applications (\Figref{fig:main}C).
Furthermore, we also identify the exposure bias issue in discrete diffusion for sequence learning~\citep{ranzato2015sequence,bengio2015scheduled}, and mitigate this by proposing a self-mixup strategy inspired by  scheduled sampling, which improves both generation quality and diversity (see \S\ref{sec:self-mixup}).

\subsection{Efficient Warm-up from Pre-trained Sequence-based DPLM}
\label{sec:method-train}
Protein sequences encode critical evolutionary information, reflecting co-evolutionary processes where residue pairs mutate together and often interact in 3D space, offering insights for predicting protein folding~\citep{melnyk2022alphafold}. \citet{lin2022esmfold} further showed that protein language models trained on large-scale evolutionary data implicitly capture this information, which can facilitate structure prediction.
Motivated by the link between evolutionary knowledge and structural interactions, we propose to built \method with an efficient warmup from pre-trained sequence-based DPLM, to make the most of established evolutionary information for protein structure modeling, 
Since our structure dataset is significantly smaller than UniRef50 sequence database (200K \vs 45M), 
enabling efficient fine-tuning of the pre-trained model.
we want to keep the sequence knowledge intact and reduce the risk of catastrophic forgetting, we apply LoRA~\citep{hu2021lora} to limit too much deviation to the original parameters.
This approach not only lowers training costs compared to starting from scratch but also effectively transfers valuable evolutionary information.

\subsection{Learning Structure Tokenization}
\label{sec:method-tokenizer}
The core difficulty of achieving a mulimodal protein LM lies in enabling the language model to learn structural information, which is challenging and remains elusive, 
Tokenizing continuous data modalities into discrete representations~\citep{van2017vqvae} has gained attraction across domains like image synthesis due to its ability to capture compact, meaningful information, enabling effective compression and efficient generation, especially with sequence-based models like Transformers.
Recent efforts have applied this approach to protein structure coordinates~\citep{van2024foldseek,haiyan2023diffusion,gao2024foldtoken,lu2024tokenized}. 
This allows language models to better learn the composition of local structural elements. However, how to learn an effective structure tokenizer remains an active research question.

\clearpage

\begin{wrapfigure}[9]{r}{0.5\textwidth}
    \centering
    \vspace{-1mm}
    \includegraphics[width=0.5\textwidth]{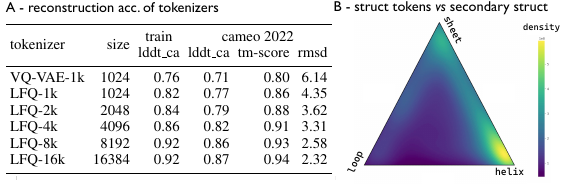}
    \vspace{-6.5mm}
    \caption{Reconstruction and secondary structure correspondence of structure tokenizers.}
    \label{fig:tokenizer}
\end{wrapfigure}
Structure tokenization under a typical VQ-VAE~\citep{van2017vqvae} framework can be summarized as follows:
\[\mathbf{x} \xrightarrow{\text{encoder}} \mathbf{e} \xrightarrow{\text{quantizer}}\mathbf{z} \xrightarrow{\text{decoder}} \tilde{\mathbf{x}}, 
\]
where (1) a structure encoder encodes backbone 3D coordinates \(\mathbf{x} \in \mathbb{R}^{L \times N_\text{backb} \times 3}\) into invariant features $\mathbf{e} \in \mathbb{R}^{L \times d_\text{quant}} $, (2) a quantizer converts $\mathbf{e}$ into $\mathbf{z}$ of $L$ discrete tokens where \(z_i \in \{0, 1, \ldots, |\mathcal{Z}|\}\) given a finite-size codebook $\mathcal{Z}$; and (3) a structure decoder reconstructs 3D coordinates $\tilde{\bm{x}}$ from the discrete tokens. 
We utilize a GVP-based~\citep{jing2020gvp} structure encoder from pre-trained GVP-Transformer~\citep{hsu2022esmif} and a IPA-based~\citep{jumper2021AF2} structure decoder.  
In terms of quantizer, our preliminary experiment showed that conventional VQ-VAE pretty much struggles in training. 
To mitigate this, we instead adopts Lookup-Free Quantizer (LFQ) from the currently best visual tokenizer~\citep{yu2023language} to protein structure tokenization.
Specifically, the latent space of LFQ is decomposed as the Cartesian product of single-dimensional binary variables, as $\mathbb{C} = \times_{k=1}^{\log_2 |\mathcal{Z}|} \mathcal{C}_k$, where $\mathcal{C}_k = \{-1, 1\}$.
Given the encoded feature $\mathbf{e} = \text{encoder}(\mathbf{x}) \in \mathbb{R}^{L \times \log_2 |\mathcal{Z}|}$, each dimension (indexed by $k$) of the quantized representation $\mathtt{quant}(e_i)$ is obtained from:
\begin{equation}
    \mathtt{quant}(e_i)[k] = \mathcal{C}_{i,k} = \mathtt{sign}(e_i[k]) = -\mathbf{1} \{z_i[k] \leq 0\} + \mathbf{1} \{e_i[k] > 0\}. \nonumber
\end{equation}
As such, with LFQ, the token indices for $\mathbf{z} = \{z_1,z_2,..., z_i,..., z_L\}$ is given by:
\begin{equation}
    z_i = \mathtt{index}(\mathtt{quant}(e_i)) = \textstyle\sum_{k=1}^{\log_2 |\mathcal{Z}|} 2^{k-1} \mathbf{1}\{e_i[k] > 0\},~ \forall z_i \in \mathbf{z}. \nonumber
\end{equation}
The LFQ-based structure tokenizer is trained on the same structure dataset as mentioned before, using a combination of reconstruction, commitment, and entropy regularization losses, similar to standard VQ-VAE. 
Here FAPE loss~\citep{jumper2021AF2} is used as the primary reconstruction loss. 


\paragraph{Evaluation.} 
As shown in \Figref{fig:tokenizer}A, LFQ significantly outperforms VQ-VAE regarding reconstruction accuracy while training of LFQ is much faster than VQ-VAE (2 \vs 15 days on 8 A100s).
Increasing codebook size leads to improved reconstruction while a codebook size of 8192 achieves the best compression-reconstruction trade-off.
Meanwhile in \Figref{fig:tokenizer}B, we observe a strong correlation between structure tokens and secondary structures. For instance, a lot of structure tokens concentrated at the alpha helix and beta sheet vertices, while some tokens lie between regions. This suggests that structure tokens the fine-grained structural elements in backbone local environment. 

\section{Experiments}
\label{sec:experiment}

In this section, we evaluate \method on various generative and understanding scenarios, including unconditional protein generation~(structure, sequence, and structure-sequence co-generation, \Secref{sec:exp-uncond}), and a variety of conditional tasks, such as folding~(\Secref{sec:exp-folding}), inverse folding~(\Secref{sec:exp-invfold}) and motif-scaffolding (\Secref{sec:exp-motif}), and a series of protein predictive tasks (\Secref{sec:exp-repr}).

\begin{figure*}[h]
    \vspace{-3mm}
    \centering
    \includegraphics[width=\linewidth]{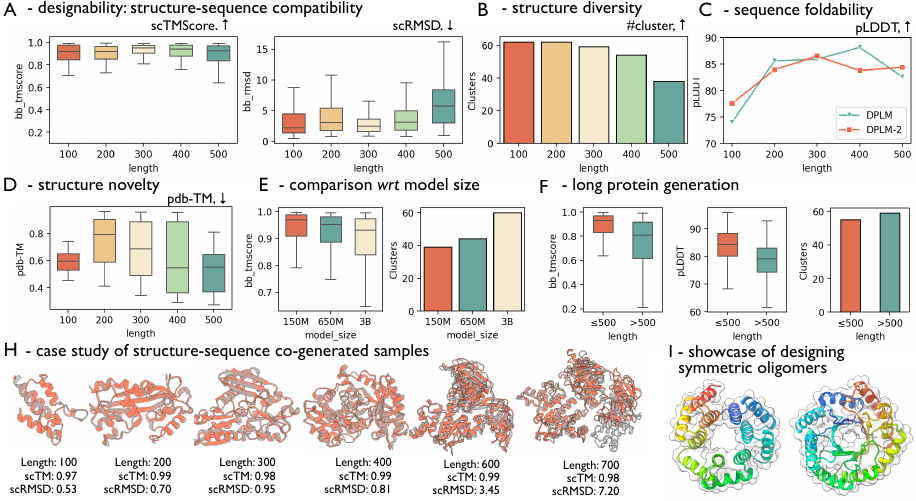}
    \vspace{-3mm}
    \caption{{\sl Evaluation of \method on unconditional structure-sequence co-generation.}
    Here for designability of co-generated proteins, we use ESMFold to obtain refolded structure of \method-generated sequence and measure the structural similarity between \method-generated structure and the refolded structure, which aims to measure the compatibility of the co-generated structure and sequence pairs.
    }
    \label{fig:uncond_all}
    \vspace{1mm}
\end{figure*}

\subsection{Unconditional Protein Generation}
\label{sec:exp-uncond}
The goal of unconditional protein generation is to produce both the 3D structure and amino acid sequence. Typically, this is done using a cascaded approach: either generating the structure first and then use another model to predict the sequence, or vice versa. Here, we focus on generating structure and sequence simultaneously. We evaluate \method on both cascaded and simultaneous generation across three tasks: \textit{unconditional structure generation}, \textit{unconditional sequence generation}, and \textit{structure-sequence co-generation}.

Following Multiflow~\citep{campbell2024generative}, we evaluate the generated proteins in terms of \textit{quality}, \textit{novelty} and \textit{diversity}.
\textbf{Quality} is measured through \textit{designability} (structure's ability to fold into a valid sequence) and \textit{foldability} (sequence's ability to fold into a reasonable structure). Designability is assessed by folding the generated sequence with ESMFold~\citep{lin2022esmfold}, then using \metric{sc-TMscore} and \metric{sc-RMSD} with the co-generated structure to evaluate similarity. 
Foldability is evaluated via ESMFold, with \metric{pLDDT} $>$ 70 considered plausible.
\textbf{Novelty} is assessed by comparing generated structures to known ones in PDB using TMScore (\metric{pdb-TM}), with lower values indicating greater novelty.
\textbf{Diversity} is measured by calculating pairwise \metric{TMscore} (\metric{inner-TM}), where lower scores indicate more dissimilarity. 
The number of clusters identified by FoldSeek~\citep{van2023foldseek} also quantifies diversity, normalized by the total number of structures.

\subsubsection{\method Enables High-quality, Diverse and Novel Protein Sequence and Structure Generation}
\label{subsec:main results}

\Tabref{tab:uncond_main} and \Figref{fig:uncond_all} 
present the results of \method for unconditional protein generation.
We highlight our key findings in the following aspects:

\textbf{(1) \method can generate diverse and highly-plausible protein with simultaneous structure-sequence co-generation.} 
We sampled 100 proteins for each length in 100, 200, 300, 400, and 500.
\Figref{fig:uncond_all}A/B demonstrates that \method can sample sequence and structures with high designability across various lengths, with most \metric{sc-TM} values exceeding 0.9, with diverse structure clusters.
\Figref{fig:uncond_all}D shows that the novelty of sampled proteins, measured by \metric{pdb-TM}, generally increases with longer protein lengths.
In addition, \method can generate with both modalities simultaneously or a modality-by-modality.
As shown in \Tabref{tab:uncond_main}, the co-generation performance exhibit highest \metric{scTM}, suggesting that co-modeling indeed benefits protein generation. 




\textbf{(2) \method can attains competitive performance with strong baselines on co-generation, as well as backbone-only and sequence-only generation, respectively.}
As shown in \Tabref{tab:uncond_main}, \method achieves the strong \metric{sc-TM} compared to strong baselines, approaching the quality of native structures from PDB. 
We notice that ESM3-Open~\citep{hayes2024esm3}, which runs in a sequence-then-structure order, fails short of unconditional generation.
Compared to MultiFlow~\citep{campbell2024generative}, \method achieves comparable co-generation quality.
Notably, as also reported in \citet{campbell2024generative}, Multiflow falls short of sequence generation when directly trained from structures with native sequences, resulting in greatly degraded co-generation performance without data distillation from external inverse folding models (ProteinMPNN). 
For reference, we also provide the result of Multiflow retrained using our training data, where its co-generation performance remains unsatisfying and lags behind \method, which suggests that \method has advantages of directly and effectively learning from complex structure-sequence joint distribution.
Moreover, \method can also only produce single modality if needed, where it matches the best competitive models in these settings respectively.
These results demonstrate \method's effectiveness as a mulitmodal generative model.

\textbf{(3) \method generates longer proteins beyond training data.}
As \method is trained with a $512$ length cutoff, we are curious about its length extrapolation, and evaluate sampled proteins at lengths of $[600,700,800, 900,1000]$.
As shown in \Figref{fig:uncond_all}F, notably, for proteins exceeding the maximum training length of 512, the \metric{pLDDT} scores of sequences sampled by \method are close to those of DPLM.
This suggests that \method largely retains its sequence generation capability inherited from sequence pre-training in DPLM, leading to its capability of length extrapolation.

\textbf{(4) Case study.}
\Figref{fig:uncond_all}H shows some generated samples of \method up to 700 residues, while in \Figref{fig:uncond_all}I we showcase that we can manipulate \method to design symmetric oligomers by forcing to duplicate the predicted tokens with repetitive structure and sequence patterns.

\begin{table*}[h]
    \centering
    \vspace{4mm}
    \setlength{\tabcolsep}{2.5pt}
    \caption{{\sl Benchmarking comparison of unconditional protein generation, in terms of structure-sequence co-generation, backbone-only generation, and sequence-only generation.} 
    For each method, we generate $100$ samples for lengths in $[100, 200, 300, 400, 500]$.
    * denotes Multiflow variants retrained by us using different dataset -- native PDB data without ProteinMPNN distillation and the same training data as \method (\ie, PDB+SwissProt), respectively.
    }
    \resizebox{\textwidth}{!}{
    \begin{tabular}{lcccccc}
        \toprule
          \multicolumn{1}{c}{} & \multicolumn{3}{c}{\textbf{Quality}}  & \multicolumn{1}{c}{\textbf{Novelty}} & \multicolumn{2}{c}{\textbf{Diversity}} \\
          \cmidrule(lr){2-4} \cmidrule(lr){5-5} \cmidrule(lr){6-7}
          
        & scTM ($\uparrow$) & scRMSD ($\downarrow$) & pLDDT ($\uparrow$) & avg. pdb-TM ($\downarrow$) & avg. inner-TM ($\downarrow$) & MaxCluster ($\uparrow$)\\

         \midrule
         \ul{\textbf{Structure-sequence co-generation.}} & & & & & &  \\
         Native PDB protein         & 4.623 $\pm$ 5.688 & 0.904 $\pm$ 0.129 & -- & -- & -- & -- \\
         ESM3-Open (1.4B, seq $\rightarrow$ struct)     
         & 0.624 $\pm$ 0.232 & 24.180 $\pm$ 24.109 & -- & 0.660 $\pm$ 0.000 & 0.410 $\pm$ 0.167 & 0.540 \\
         
         MultiFlow \textit{w/} distillation (official ckpt) &\textbf{0.930} $\pm$ 0.098 & 3.208 $\pm$ 4.741 & 79.447 & 0.704 $\pm$ 0.000 & 0.468 $\pm$ 0.152 & 0.500 \\
         *MultiFlow \textit{w/o} distillation & 0.750 $\pm$ 0.163 & 9.306 $\pm$ 8.499 & 65.861 \\
         *MultiFlow (retrained on our training data) & 0.871 $\pm$ 0.934 & 6.580 $\pm$ 6.258 & 62.624   \\
         \chl \method (650M, seq $\rightarrow$ struct) 
         & \chl 0.907 $\pm$ 0.117 & \chl 6.337 $\pm$ 9.403 & \chl 82.246 & \chl 0.653 $\pm$ 0.195 & \chl 0.594 $\pm$ 0.270 & \chl \textbf{0.651} \\
         \chl \method (650M, struct $\rightarrow$ seq) 
         & \chl 0.921 $\pm$ 0.098 & \chl 4.969 $\pm$ 6.735 & \chl 81.910 & \chl 0.637 $\pm$ 0.195 & \chl 0.679 $\pm$ 0.288 & \chl 0.575 \\
          \chl \textbf{\method (650M, co-generation)} 
         & \chl \textbf{0.925} $\pm$ 0.085 & \chl 3.899 $\pm$ 3.723 & \chl 82.686 & \chl 0.640 $\pm$ 0.204 & \chl 0.703 $\pm$ 0.279 & \chl 0.545 \\

        \midrule
         \multicolumn{6}{l}{\ul{\textbf{Unconditional backbone generation.} (sequence predicted by ProteinMPNN)}} \\
         Native PDB struct.~(seq. from PMPNN)            & 0.969 $\pm$ 0.000 & 0.864 $\pm$ 0.000 & -- & -- & 0.282 $\pm$ 0.000 & 0.782 \\
         FrameDiff               & 0.818 $\pm$ 0.000 & 3.919 $\pm$ 0.000 & -- & 0.668 $\pm$ 0.000 & 0.465 $\pm$ 0.000 & 0.252 \\
         FoldFlow                & 0.540 $\pm$ 0.000 & 7.965 $\pm$ 0.000 & -- & 0.566 $\pm$ 0.000 & 0.411 $\pm$ 0.000 & 0.762 \\
         RFDiffusion             & 0.914 $\pm$ 0.000 & 1.969 $\pm$ 0.000 & -- & 0.657 $\pm$ 0.000 & 0.363 $\pm$ 0.000 & 0.598 \\

         \chl \textbf{\method} (650M)                & \chl \textbf{0.945} $\pm$ 0.082 & \chl 4.451 $\pm$ 5.261 & \chl -- & \chl 0.637 $\pm$ 0.195 & \chl 0.679 $\pm$ 0.288 & \chl 0.575 \\

         \midrule
         \multicolumn{6}{l}{
         \ul{\textbf{Unconditional sequence generation.} (structures predicted by ESMFold)}}  \\
         EvoDiff 
         & -- & -- & 35.846 & 0.432 $\pm$ 0.106 & 0.366 $\pm$ 0.070 & 0.990 \\
         DPLM (650M)         
         & -- & -- & 83.252 & 0.541 $\pm$ 0.187 & 0.515 $\pm$ 0.222 & 0.735 \\
         \chl \textbf{\method} (650M)
         & \chl -- & \chl -- & \chl 82.246 & \chl 0.662 $\pm$ 0.199 & \chl 0.589 $\pm$ 0.268 & \chl 0.700 \\
        \bottomrule
        \end{tabular}
        }

    \label{tab:uncond_main}
\end{table*}

\begin{figure*}[h]
    \vspace{3mm}
    \centering
    \includegraphics[width=\linewidth]{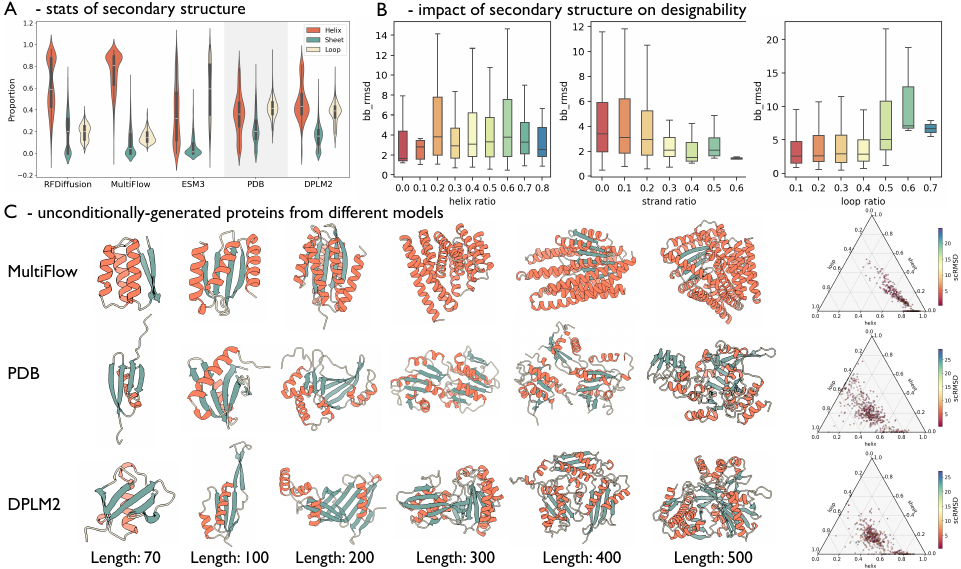}
    \vspace{-1mm}
    \caption{{\sl Analysis regarding secondary structure of generated proteins.}
    \textbf{(A)} Statistics of averaged proportions of secondary structures for proteins from different methods and PDB; 
    \textbf{(B)} Secondary structure \vs designability; 
    \textbf{(C)} Samples of Multiflow, PDB and DPLM-2, as well as their secondary structure distributions.
    }
    \label{fig:ssp_all}
    \vspace{1mm}
\end{figure*}


\subsubsection{\method Generates Proteins That Resembles Natural Proteins}

To further analyze the properties of different model, we examine their secondary structure distribution against natural proteins from PDB.

\paragraph{Proteins sampled by \method have secondary structures most similar to natural proteins.} As seen in \Figref{fig:ssp_all}A, structure-based models like RFDiffusion and MultiFlow generate proteins with more helices and fewer sheets and loops than natural proteins in PDB. Protein language models like ESM3 and \method show no strong bias towards alpha helices, but ESM3 tends to generate more loops. Among the methods, \method produces the most natural-like secondary structure proportions, closely matching PDB proteins.
In \Figref{fig:ssp_all}C, proteins generated by MultiFlow contain many helices and become more globular as length increases, exhibiting idealized secondary structures. 
In contrast, proteins generated from \method resembles natural ones have more balanced structures, with fewer helices and more beta sheets and loops.
On the other hands, simplex plots in \Figref{fig:ssp_all}C shows that while MultiFlow's proteins are clustered in helix-rich regions, \method's proteins span a wider area similar to natural proteins, while it rarely samples proteins composed mostly of sheets and loops, which do occur in nature.
Additionally, \Figref{fig:ssp_all}B shows that the loop ratio has a significant impact on designability, where a higher proportion of loops will increase \metric{scRMSD}, as loops are highly flexible. Thus, proteins with long loops, which \method often generates, tend to have relatively high \metric{scRMSD}, aligning with the results in \Tabref{tab:uncond_main}.

\subsubsection{Ablation Study}

In \method training, we start with a warmup from the sequence-based pre-trained DPLM to exploit established evolutionary information and augment the data with high-quality AlphaFold-predicted structures from SwissProt (around 200K) and clustered PDB structures. This section evaluates the effects of sequence pre-training and data augmentation on unconditional protein generation.

\begin{table}[h]
\centering
\vspace{1mm}
\setlength{\tabcolsep}{8pt}
\caption{Ablation study on the sequence pre-training and training data augmentation.}
\small
\resizebox{\linewidth}{!}{%
\begin{tabular}{cccccccccccc}
\toprule
 \multirow{2}{*}{\shortstack{sequence\\pre-training}} 
 & \multirow{2}{*}{\shortstack{synthetic\\structures}} 
 & \multicolumn{2}{c}{length 100} 
 & \multicolumn{2}{c}{length 200}
 & \multicolumn{2}{c}{length 300}
 & \multicolumn{2}{c}{length 400}
 & \multicolumn{2}{c}{length 500} \\
 \cmidrule[0.3pt](lr){3-4}
 \cmidrule[0.3pt](lr){5-6}
 \cmidrule[0.3pt](lr){7-8}
 \cmidrule[0.3pt](lr){9-10}
 \cmidrule[0.3pt](lr){11-12}
 & & scTM & clusters & scTM & clusters & scTM & clusters & scTM & clusters & scTM & clusters
 \\
\midrule

 \ding{55} & \ding{55}
 & 0.9241  & 20 
 & 0.8674  & 34 
 & 0.7667  & 33 
 & 0.5016  & 25 
 & 0.4511  & 25 \\
 
 \checkmark & \ding{55}   
 & \textbf{0.9610}  & 26
 & 0.9349  & \textbf{47}
 & 0.9169  & 38
 & 0.8643  & 35
 & 0.7673  & \textbf{52} \\

 \ding{55} & \checkmark    
 & 0.8988  & 27
 & 0.9182  & 15
 & \textbf{0.9343}  & 13
 & 0.8518  & 21
 & 0.8288  & 31 \\

 \checkmark & \checkmark  
 & 0.9348  & \textbf{35} 
 & \textbf{0.9428}  & 40
 & 0.9232  & \textbf{48}
 & \textbf{0.9260} & \textbf{40}
 & \textbf{0.9012}  & 32 \\
 
\bottomrule
\end{tabular}
}
\label{tab:ablation}
\vspace{5pt}
\end{table}

\Tabref{tab:ablation} demonstrates that \textit{sequence pre-training and data augmentation can significantly improve the designability and diversity}, especially in generating long proteins~(length $> 300$).
We hypothesize that the limited number of long proteins in PDB leads to insufficient training. In contrast, sequence pretraining, which includes evolutionary data, is essential and can be transferred to improve protein structure modeling and generation quality. Additionally, this evolutionary information boosts sampling diversity. While increasing the amount of training data improves designability, it is less effective in enhancing diversity compared to sequence pretraining. By combining both strategies, we achieve the best overall performance, which forms the core of our training strategy.





\subsection{Forward Folding (Sequence-conditioned Structure Prediction)}
\label{sec:exp-folding}


\begin{wraptable}[16]{r}{0.55\linewidth}
   \centering
   \small
   \setlength{\tabcolsep}{2.5pt}
   \vspace{-1mm}
   \caption{{\sl Structure prediction performance comparison between \method and different baseline approaches on CAMEO 2022 datasets.}
   $\dagger$: PVQD results are quoted from~\citet{haiyan2023diffusion}.
   }
   \label{tab:folding}

   \resizebox{1.0\linewidth}{!}{%
   \begin{tabular}{lcccc}
   \toprule
     \multirow{2}{*}{ Models} 
    & \multicolumn{2}{c}{CAMEO 2022} 
    & \multicolumn{2}{c}{PDB date split} \\
    
    \cmidrule[0.3pt](lr){2-3} \cmidrule[0.3pt](lr){4-5}
      & {\metric{RMSD}} & {\metric{TMscore}} & {\metric{RMSD}} & {\metric{TMscore}} \\
   \midrule

    ESMFold 
   & 3.99/2.03  &  0.85/0.93    
   & 2.84/1.19  &  0.93/0.97    
   \\
    $^\dagger$PVQD 
   & 4.08/1.95  &  0.81/0.88   
   & -- & --
   \\
    MultiFlow 
   & 17.84/17.96  &  0.50/0.46  
   & 15.64/16.08  &  0.53/0.49
   \\
    ESM3
   & 6.33/2.98  &  0.85/0.92   
   & 4.94/2.28  &  0.87/0.93
   \\
  \midrule
   \method~(150M)
  &  9.22/7.64  &  0.75/0.81  
  &  8.35/5.60  &  0.76/0.82  \\ 
  
    \quad \textit{w/} folding SFT 
  &  7.66/4.37  &  0.80/0.86    
  &  6.00/3.41  &  0.83/0.88 \\ 
  
  \method~(650M)   
  &  7.37/4.89  &  0.79/0.86   
  &  5.67/3.33  &  0.83/0.88 \\ 
  
   \quad \textit{w/} folding SFT 
  &  6.21/3.78  &  0.84/0.89    
  &  3.40/1.78  &  0.89/0.94 \\ 
  
   \chl \method~(3B)     
  & \chl 6.34/3.65  & \chl 0.83/0.89   
  & \chl 4.54/2.54  & \chl 0.86/0.92 \\ 
  
   \chl \quad \textit{w/} folding SFT 
  & \chl 5.71/3.23  & \chl 0.85/0.90    
  & \chl 3.15/1.69  & \chl 0.90/0.95 \\ 
  
\bottomrule
\end{tabular}
}
   \vspace{8pt}
\end{wraptable}

The goal of folding is to predict the 3D structure for the given amino acid sequence~\citep{jumper2021AF2}.
As a mulitmodal generative model, \method spontaneously enables protein structure prediction task (see \Figref{fig:main}C-3) given sequence as conditioning.
We assess \method on CAMEO 2022 and a PDB data split used by Multiflow~\citep{campbell2024generative}.
We utilize \metric{RMSD} and \metric{TMscore} between predicted and ground truth structure for evaluation, while \method adopts \metric{argmax} decoding for 100 sampling iterations.

\paragraph{\Tabref{tab:folding} indicates that \method can perform sufficiently good folding in a zero-shot manner.} 
Performance can be improved after further supervised fine-tuning (SFT) using folding objective ($\max_{\theta} \log p_\theta (\bm{z} | \bm{s}) $). 
Overall, \method can outperform or on par with the strong baselines, while achieving close performance with ESMFold.
Furthermore, We observe that \method with larger model scales can attain better results than smaller ones.
We suggest that \method benefits from the evolutionary information inherited from DPLM pre-trained on the vast number of protein sequences, which can be transferred and leveraged into structure modeling. 

\subsection{Inverse Folding (Structure-conditioned Sequence Generation)}
\label{sec:exp-invfold}

The goal of inverse folding is to find an amino acid sequence that can fold to a given backbone structure.
For evaluation, we employ amino acid recovery (\metric{AAR}) for sequence evaluation, and we also assess the structure by self-consistency TM-score~(\metric{scTM}) between the native structure and the ESMFold-predicted structure of the generated sequence.

\begin{wraptable}[9]{r}{0.55\linewidth}
   \centering
   \small
   \setlength{\tabcolsep}{3pt}
   \vspace{-1mm}
   \caption{{\sl Comparison on inverse folding task.}
   }
   \label{tab:invfold}

   \resizebox{1.0\linewidth}{!}{%
   \begin{tabular}{lcccc}
   \toprule
     \multirow{2}{*}{ Models} 
    & \multicolumn{2}{c}{CAMEO 2022} 
    & \multicolumn{2}{c}{PDB date split} \\
    \cmidrule[0.3pt](lr){2-3} \cmidrule[0.3pt](lr){4-5}
   
  & \metric{AAR} & \metric{scTM} 
  & \metric{AAR} & \metric{scTM} \\
  \midrule
   {MultiFlow} 
   &  32.28/33.58  & 0.87/0.94 & 37.74/37.59 & 0.94/0.96   \\
   ESM3
   & 47.06/46.24  &  0.90/0.95   
   & 49.50/49.42  &  0.94/0.97
   \\

  \midrule
     \method~(150M)
  &  45.22/46.12  & 0.87/0.93
  &  48.83/47.96  &  0.89/0.95 \\ 
  
    \method~(650M)   
  &  49.01/50.10  & 0.88/0.93
  &  54.80/53/07  &  0.91/0.96  \\ 
  
   \chl \method~(3B)     
  & \chl 52.36/53.72  & \chl 0.89/0.95
  & \chl 61.67/57.91  & \chl 0.92/0.96  \\ 
  
\bottomrule
\end{tabular}
}
   \vspace{8pt}
\end{wraptable}

\paragraph{\method can generate reasonable sequences that fold into the given structures.} 
\Tabref{tab:invfold} presents that \method can outperform or be on par with other co-generation models~(MultiFlow, ESM3).
As the model size increases, the performance in terms of sequence recovery~(\metric{AAR}) and structural consistency~(\metric{scTM}) improves, revealing the same scaling law observed in the folding task.
We suggest that multimodal training effectively aligns the structure and sequence into the same space, such that \method can yield the corresponding sequence without additional training.

\subsection{Scaffolding with Mixed-modal Motif Conditioning}
\label{sec:exp-motif}

The objective of motif-scaffolding is to generate a suitable scaffold to preserve the structure of the given motif and maintain its original function.
We follow the experimental setting of \citet{yim2024improved}, with 24 motif-scaffolding problems and we sample 100 scaffolds for each motif, where we 
(1) first determine the length of scaffold, and then
(2) keep the motif segment unchanged and sample the scaffold part conditioned on the motif.
The scaffold length is sampled from a range provided by \citet{yim2024improved}, and when there are multiple motifs, the order of motif segments is consistent with \citet{yim2024improved}.
We provide the 3D structure and sequence of motif as input of \method.
As a multimodal model, we evaluate \method using sequence-based, structure-based, and co-generation approaches.
A scaffold is considered successful if it satisfies both criteria
(1) overall designablity, which is successful when \metric{pLDDT} $> 70$ (for sequence-based models) or \metric{scTM} $> 0.8$, and
(2) motif-preseving, which is deemed successful when the predicted motif structure matches the native one with \metric{motif-RMSD} $<$1\AA.

\begin{wrapfigure}{r}{0.35\textwidth}
    \centering
    \includegraphics[width=\linewidth]{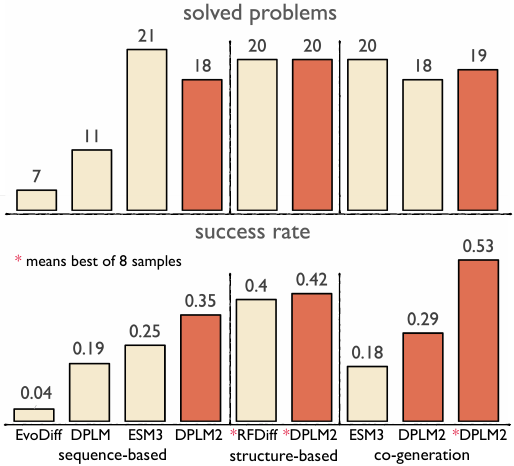}
    \vspace{-5mm}
    \caption{{\sl Evaluation of motif-scaffolding \wrt success rate and num. of solved problems.}
    }
    \label{fig:motif_main}
\end{wrapfigure}
\paragraph{\Figref{fig:motif_main} reveals that \method is capable of generate reasonable scaffolds for the given functional motifs.}
In sequence-based, structure-based and co-generation evaluation, \method  can outperform or be on par with the corresponding approaches in most cases, solving more motif problem and achieving higher average success rate. 
We compared to sequence-based method, \method shows better performance since it allows structural input of motif, which is important for preserving motif's structure hence the functions.
Remarkably, \method attains comparable performance with RFDiffusion when only generating scaffold structure, while achieve better performance when simultaneously designing scaffold sequence and structure, outperforming ESM3.
Despite not experimentally verified, these results suggest that with \method, mulitmodal conditioning and generation could lead to more successful conditional protein design.


\begin{table*}[h]
\centering
\footnotesize
\vspace{2mm}
\setlength{\tabcolsep}{4.5pt}
\caption{{\sl Performance on various protein predictive downstream tasks.}
$\dagger$: benchmarked results are quoted from \citet{su2023saprot}.
}
\label{tab:results_understanding}
\vspace{1.5pt}
\resizebox{\linewidth}{!}{%
\begin{tabular}{lccccccccc}
\toprule
\multirow{3}{*}{{Models}} & \multirow{2}{*}{Thermostability} & \multirow{2}{*}{HumanPPI} & \multirow{2}{*}{Metal Ion Binding} & \multirow{2}{*}{EC} & \multicolumn{3}{c}{GO}                           & \multicolumn{2}{c}{DeepLoc}     \\
\cmidrule[0.5pt](lr){6-8}  \cmidrule[0.5pt](lr){9-10} &  &  &  &  & MF & BP & CC & Subcellular   & Binary   \\ 
\cmidrule[0.5pt](lr){2-10} 
    & \metric{Spearman's} $\rho$      & \metric{Acc} ($\%$)       & \metric{Acc} ($\%$)    & \metric{Fmax}      & \metric{Fmax}   & \metric{Fmax}    & \metric{Fmax}    & \metric{Acc} ($\%$)   & \metric{Acc} ($\%$)    \\ 
\midrule
$^\dagger$SaProt (650M)    & 0.724    & 86.41      & 75.75    & 0.884      & 0.678    & 0.356 & 0.414    & 85.57  & 93.55   \\ 

$^\dagger$MIF-ST~\citep{yang2022masked} & 0.694 & 75.54 & 75.08 & 0.803 & 0.627 & 0.239 & 0.248 & 78.96 & 91.76  \\

\midrule
ESM2 (650M)   & 0.691   & 84.78   & 71.88     & 0.866       & 0.676 & 0.344   & 0.402  & 83.68   & 92.28  \\
 DPLM (650M) &  {0.695}           &   {86.41}     &  {75.15}                     &  {0.875}      &  {0.680} &  {0.357} &  {0.409} &  {84.56} &  {93.09}  \\
\midrule

\chl \method (650M)  & \chl \textbf{0.714}           &  \chl 84.44    & \chl {74.28}                     & \chl \textbf{0.878}      & \chl {0.680} & \chl \textbf{0.359} & \chl \textbf{0.411} & \chl {82.98} & \chl \textbf{93.64}  \\

\bottomrule
\end{tabular}
}
\vspace{2mm}
\end{table*}

\subsection{Evaluation of Protein Representation Learning}
\label{sec:exp-repr}

Directly access to structure information is supposed to benefit downstream protein predictive tasks.
To inspect this, we evaluate \method on a variety of protein predictive tasks utilizing the dataset provided by SaProt~\citep{su2023saprot}, where we provide tokenized protein structure tokens along with the protein sequences to \method.

\begin{wraptable}[9]{r}{0.25\textwidth}
\centering
\footnotesize
\vspace{-1mm}
\setlength{\tabcolsep}{4.5pt}
\caption{{\sl Performance without large-scale sequence pre-training.}
}
\label{tab:results_understanding_noptrn}
\vspace{1.5pt}
\resizebox{\linewidth}{!}{%
\begin{tabular}{lc}
\toprule
\multirow{3}{*}{{Models}}  & {DeepLoc}     \\
 &  Subcellular   \\ 
\cmidrule[0.5pt](lr){2-2} 
    & \metric{Acc} ($\%$)    \\ 

\midrule
 DPLM (650M) &  {63.49}  \\
\midrule
\chl \method (650M)  & \chl \textbf{66.77}  \\

\bottomrule
\end{tabular}
}
\end{wraptable}
\paragraph{\method can perform multimodal representation learning by leveraging both structure and sequence information.}
\Tabref{tab:results_understanding} presents that \method shows further improvement compared to sequence-only methods~(ESM2, DPLM) on some tasks, indicating that \method can leverage protein structures to generate better representations containing multimodal information for downstream tasks.
However, we find that \method falls behind the state-of-the-art structure-aware protein LM, \ie, SaProt, in most tasks and even lags behind DPLM in certain tasks.
We hypothesize this is because the strutcure training data of DPLM-2, consisting of PDB and SwissProt, is smaller and differs from UniRef50, which DPLM is pretrained on, potentially causing catastrophic forgetting and suboptimal representation. To test this, we conducted an experiment on the DeepLoc subcellular task, where \method underperforms compared to DPLM. As shown in \Tabref{tab:results_understanding_noptrn}, without large-scale sequence pretraining, \method outperforms DPLM significantly, suggesting that: (1) Incorporating structure information enhances performance over sequence-only models. (2) Smaller datasets can lead to catastrophic forgetting, diminishing the benefits of large-scale pretraining.
As result, to further improve the predictive performance, one deserving direction is to exploit larger-scale predicted structures in our future work.

%
%

\section{Discussions}
In this paper, we introduce \method, a multimodal diffusion protein language model that understands, generates and reasons over protein structure and sequence, aiming to severe as a mulimodal foundation for protein. 
Despite promising performance spanning protein co-generation, folding, inverse folding and conditional motif-scaffolding with mulimodal input and output, there remains several limitations deserving to be addressed.
(1) Structure data: Our findings indicate that while structure awareness may help with predictive tasks, the limited structure data constrains \method's ability to learn robust representations. It is also important to account for longer protein chains and multimers in future studies.
(2) Trade-off of discrete latent representation: Tokenizing structure into discrete symbols facilitates multimodal protein language models and co-generation but may come at the cost of losing fine-grained structural details and control, such as precise atomic positions and inter-atomic distances. 
Future work should aim to also integrate the strengths of data-space structure-based generative models into sequence-based mulitimodal language models to maximize the best of both worlds.


\section*{Acknowledgement}
We would like to thank Dr. Hang Li for insightful discussions on the project and feedback on the manuscript that help shape this study. 
We thank Yi Zhou, Jing Yuan, Yilai Li, Yuning Shen, Wesley Hsieh and Daiheng Zhang for their valuable comments.


\bibliography{references}
\bibliographystyle{iclr2025_conference}

\appendix

\section{\method Training}

\subsection{Tackling Exposure Bias in Discrete Diffusion with Self-mixup Training Strategy}
\label{sec:self-mixup}

We find that discrete diffusion training will face the \textit{exposure bias} problem~\citep{ranzato2015sequence, bengio2015scheduled}, which means mismatch between training and inference.
The model is trained to denoise given the ground-truth context during training.
However, during inference, the model needs to denoise based on the predicted tokens, which may not be correct and  inconsistent with the always-accurate context during training.
This may lead to error accumulation and negatively impact the generation performance.

To address this issue, we propose a \textit{self-mixup} training paradigm for discrete diffusion model, enhancing the consistency between training and inference.
During training, we perform an additional forward pass, allowing the model to first make predictions and then denoise based on those predictions.

\Tabref{tab:self-mixup} shows that the \textit{self-mixup} training strategy effectively enhances the diversity of samples. 
We attribute this to the model producing more accurate logits during inference, leading to more diverse reasonable sampling paths instead of converging on the sampling paths with the highest probability, which results in more diverse proteins.

\begin{table}[h]
\centering
\vspace{1mm}
\setlength{\tabcolsep}{8pt}
\caption{Ablation study on the \textit{self-mixup} training strategy.}
\small
\resizebox{\linewidth}{!}{%
\begin{tabular}{ccccccccccc}
\toprule
 \multirow{2}{*}{\shortstack{Mixup\\strategy}}  
 & \multicolumn{2}{c}{length 100} 
 & \multicolumn{2}{c}{length 200}
 & \multicolumn{2}{c}{length 300}
 & \multicolumn{2}{c}{length 400}
 & \multicolumn{2}{c}{length 500} \\
 \cmidrule[0.3pt](lr){2-3}
 \cmidrule[0.3pt](lr){4-5}
 \cmidrule[0.3pt](lr){6-7}
 \cmidrule[0.3pt](lr){8-9}
 \cmidrule[0.3pt](lr){10-11}
 & scTM & clusters & scTM & clusters & scTM & clusters & scTM & clusters & scTM & clusters
 \\
\midrule

 \ding{55} 
 & \textbf{0.9237}  & 44 
 & \textbf{0.9180}  & 53 
 & 0.9147  & 48 
 & 0.9059  & 42 
 & \textbf{0.8896}  & 33 \\
 
 \checkmark  
 & 0.8812  & \textbf{62}
 & 0.8820  & \textbf{62}
 & \textbf{0.9172}  & \textbf{59}
 & \textbf{0.9099}  & \textbf{54}
 & 0.8845  & \textbf{38} \\
 
\bottomrule
\end{tabular}
}
\label{tab:self-mixup}
\vspace{5pt}
\end{table}

\subsection{Dataset}
The training set of \method is composed by experimental data, \ie, PDB~\citep{berman2000protein}, and high quality synthetic data, \ie, SwissProt~\citep{varadi2022alphafold}.
We filter the SwissProt data by pLDDT $>$ 85.
After filtering, the overall training set contains approximately 200,000 proteins.
We limit the maximum length of the training set to 512.
For proteins longer than 512, we randomly crop it to 512.
We crop the low pLDDT (pLDDT $<$ 50) segments located at the both ends of proteins in the SwissProt dataset. 
These segments are typically non-structural and may negatively impact the training results.
Moreover, we find that the length distribution of the training set is not balanced, where the number of proteins with length less than 100 is relatively small, leading to a suboptimal diversity among the short proteins.
Therefore, during training, we randomly crop long proteins to short proteins with a probability of 50\% for each batch to improve the diversity.

\subsection{Hyperparameter}

We train all models using AdamW optimizer~\citep{kingma2014adam} with $\beta_1$ = 0.9 and $\beta_2$ = 0.95.
We use a weight decay of 0.01 and gradient clipping of 0.5. 
We employ 2K warmup steps until reaching the maximum learning rate, and utilize a linear decay scheduler to decay LR to 10\% of the maximum learning rate by the end of training.
The maximum learning rate is 1e-4, and the overall training step is 100,000.
We utilize the pretrained DPLM as the parameter initialization, and the diffusion timestep is set to 500.
We train 150M \method with 8 A100 GPUs for 3 days, while 650M with 16 A100 GPUs for 3 days and 3B with 16 A100 GPUs for a week.

\section{Structure Tokenizer}
The core difficulty of achieving a mulimodal protein LM lies in enabling the language model to learn structural information, which is challenging and remains elusive, 
Tokenizing continuous data modalities into discrete representations~\citep{van2017vqvae} has gained attraction across domains like image synthesis due to its ability to capture compact, meaningful information, enabling effective compression and efficient generation, especially with sequence-based models like Transformers.
Recent efforts have applied this approach to protein structure coordinates~\citep{van2024foldseek,haiyan2023diffusion,gao2024foldtoken,lu2024tokenized}. 

\subsection{Dataset}
Our structure tokenizers are trained using the same structure data as our mulitmodal language model, containing both experimental and high-quality structures, totaling 200K proteins.

\subsection{Model Architecture}
As shown in \Figref{fig:main}A, the structure tokenizer in this paper consists of a structure encoder, quantizer, and structure decoder. The encoder is based on a pre-trained GVP-Transformer~\citep{hsu2022esmif}, with its parameters frozen during training. It transforms backbone structures into geometric features, which are projected onto a latent embedding using an MLP layer. For the quantizer, we adopt a lookup-free quantizer from a state-of-the-art video tokenizer~\citep{yu2023language}, where the latent dimension is set to $\log_2 |\mathcal{Z}|$, with $|\mathcal{Z}|$ as the codebook size. The structure decoder follows the IPA-based modules from AlphaFold2~\citep{jumper2021AF2}, using 4 EvoFormer layers without MSA row attention, following ESMFold~\citep{lin2022esmfold}, to generate atomic positions from the structure tokens.

\subsection{Training}
The structure tokenizer is trained using a standard VQ-VAE framework, with the objective including reconstruction loss, codebook commitment loss, and entropy regularization loss to ensure effective codebook utilization. For the reconstruction loss, we adopt the FAPE loss, violation loss, and distogram loss from AlphaFold2, measuring the difference between predicted and native structures. To further enhance the training, we introduce a sequence prediction head on top of the structure decoder’s final representation and minimize the cross-entropy against the native sequence.

\section{Motif Scaffolding}
\subsection{Evaluation Pipeline}
\label{sec:motif_evaluation}

\begin{figure}[t]
    \centering
    \caption{{ Sequence-based, structure-based and co-generation evaluation pipeline of motif-scaffolding.} 
    }
    \includegraphics[width=0.8\linewidth]{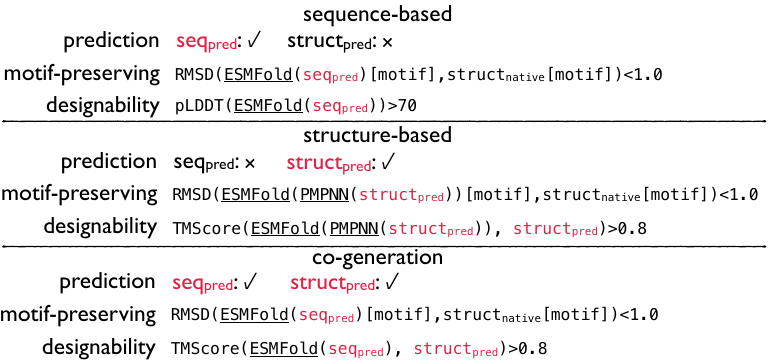}
    \vspace{5mm}
    
    \label{fig:motif_evaluation}
\end{figure}

We evaluate \method in sequence-based, structure-based and co-generation ways. The overall illustration is shown in \Figref{fig:motif_evaluation}.

We focus on the two aspects: overall quality and motif part consistency. 
The assessment of overall quality varies across different approaches. Specifically,
(1) For sequence-based method, we only take the generated sequence and utilize ESMFold to obtain the predicted structure, and the \metric{pLDDT} score provided by ESMFold is used to assess overall quality.
(2) For structure-based method, we only take the generated structure, and then leverage ProteinMPNN to predict the sequence, followed by ESMFold to predict the structure, where overall quality is assessed by \metric{scTM}.
(3) For co-generation method, we take both the generated structure and sequence, and predict structure given generated sequence with ESMFold, where \metric{scTM} is calculated between generated structure and ESMFold predicted structure to evaluate overall quality.
Considering that the ground truth motif structure is given, we only utilize the ESMFold predicted structure to calculate \metric{motif-RMSD}.

\subsection{Result of Each Problem}

\Tabref{tab:motif_each_problem} presents the result of each motif-scaffolding problem.
\method achieves the best average success rate in each evaluation.
Compared with ESM3, \method shows better results in 12 problems in co-generation evaluation and 10 problems in sequence-based evaluation.
Meanwhile, \method outperforms RFDiffusion in 14 problems in structure-based evaluation.
This demonstrates that \method can achieve strong performance under various evaluation methods.

We also find that taking the best result from 8 samples can bring significant improvement compared to 1 sample, especially in terms of success rate.
In the co-generation evaluation, DPLM2 with sampling 8 times improves the success rate of most of the problems by a large margin.
We hypothesize that sampling eight times largely alleviates errors caused by randomness in the sampling process, thereby producing a more suitable scaffold for the given motif.

\begin{table}[t]
\centering
\vspace{-1.5mm}
\caption{Motif-scaffolding results of each problem. * means best result from 8 samples. }
\label{tab:motif_each_problem}
\vspace{1.5pt}
\resizebox{0.9\linewidth}{!}{%
\begin{tabular}{cccccccccc}
\toprule
\multirow{2}{*}{} & \multicolumn{4}{c}{{sequence-based}} & \multicolumn{2}{c}{{structure-based}} 
& \multicolumn{3}{c}{{co-generation}}\\ 
\cmidrule[0.5pt](lr){2-5} \cmidrule[0.5pt](lr){6-7} 
\cmidrule[0.5pt](lr){8-10} 
& EvoDiff & DPLM & ESM3 & DPLM2
& *RFDiffusion & *DPLM2 
& ESM3 & DPLM2 & *DPLM2
\\
\midrule
1BCF  
& 0.00 & 0.00  & \textbf{0.89} & 0.01 
& \textbf{1.00} & 0.07
& \textbf{0.23} & 0.01  & 0.05 \\
1PRW  
& 0.61 & 0.83  & \textbf{0.96} & 0.86
& 0.08 & \textbf{0.96}
& 0.54 & 0.84  & \textbf{0.95} \\
1QJG  
& 0.00 & 0.00  & 0.02 & \textbf{0.03}
& 0.00 & 0.00
& 0.03 & 0.02  & \textbf{0.05} \\
1YCR 
& 0.02 & 0.38  & 0.41 & \textbf{0.77}
& 0.74 & \textbf{0.93}
& 0.18 & 0.53  & \textbf{0.98} \\
2KL8
& 0.04 & 0.08  & 0.11 & \textbf{0.47}
& 0.88 & \textbf{0.94}
& 0.11 & 0.57  & \textbf{1.00} \\
3IXT
& 0.06 & 0.17  & 0.18 & \textbf{0.67}
& 0.25 & \textbf{0.77}
& 0.02 & 0.41  & \textbf{0.73} \\
4JHW
& 0.00 & 0.00  & 0.00 & 0.00
& 0.00 & 0.00
& 0.00 & 0.00  & 0.00 \\
4ZYP
& 0.00 & 0.00  & 0.03 & \textbf{0.16}
& 0.40 & \textbf{0.51}
& 0.08 & 0.10  & \textbf{0.64 }\\
5IUS
& 0.00 & 0.00  & 0.00 & 0.00
& \textbf{0.02} & 0.00
& 0.00 & 0.00  & 0.00 \\
5TPN
& 0.00 & 0.00  & \textbf{0.03} & 0.00
& \textbf{0.61} & 0.06
& \textbf{0.01} & 0.00  & 0.00 \\
5TRV\_long
& 0.00 & 0.00  & \textbf{0.19} & 0.00
& \textbf{0.37} & 0.08
& \textbf{0.19} & 0.00  & 0.07 \\
5TRV\_med
& 0.00 & 0.00  & \textbf{0.16} & 0.03
& \textbf{0.24} & 0.07
& 0.16 & 0.02  & \textbf{0.19} \\
5TRV\_short
& 0.00 & 0.00  & 0.01 & \textbf{0.07}
& 0.04 & \textbf{0.10}
& 0.01 & 0.03  & \textbf{0.11} \\
5WN9
& 0.00 & 0.00  & \textbf{0.02} & 0.00
& 0.00 & \textbf{0.20}
& 0.00 & 0.00  & 0.00 \\
5YUI              
& 0.00 & 0.00  & 0.00 & 0.00
& \textbf{0.02} & 0.00
& 0.00 & 0.00  & 0.00 \\
6E6R\_long
& 0.01 & 0.65  & 0.07 & \textbf{0.91}
& 0.86 & \textbf{0.92}
& 0.04 & 0.78  & \textbf{1.00} \\
6E6R\_med
& 0.03 & \textbf{0.94}  & 0.24 & 0.93
& \textbf{0.89} & 0.88
& 0.14 & 0.77  & \textbf{0.97} \\
6E6R\_short
& 0.07 & \textbf{0.87}  & 0.09 & 0.86
& 0.39 & \textbf{0.78}
& 0.06 & 0.64  & \textbf{0.99} \\
6EXZ\_long
& 0.00 & 0.01  & 0.32 & \textbf{0.61}
& \textbf{0.76} & 0.63
& 0.13 & 0.44  & \textbf{0.95} \\
6EXZ\_med
& 0.00 & 0.00  & 0.31 & \textbf{0.66}
& 0.49 & \textbf{0.63}
& 0.31 & 0.55  & \textbf{0.96} \\
6EXZ\_short
& 0.00 & 0.00  & 0.31 & \textbf{0.66}
& 0.39 & \textbf{0.41}
& 0.28 & 0.58  & \textbf{0.87} \\
7MRX\_long
& 0.00 & 0.02  & \textbf{0.36} & 0.23
& 0.09 & \textbf{0.32}
& 0.37 & 0.20  & \textbf{0.73} \\
7MRX\_med
& 0.00 & 0.31  & \textbf{0.65} & 0.28
& 0.11 & \textbf{0.31}
& 0.59 & 0.22  & \textbf{0.70} \\
7MRX\_short
& 0.00 & 0.34  & \textbf{0.68} & 0.26
& 0.02 & \textbf{0.41}
& 0.74 & 0.24  & \textbf{0.88} \\
\midrule
pass rate         
& 7/24 & 11/24  & \textbf{21/24} & 18/24
& 20/24 & 20/24
& \textbf{20/24} & 18/24  & 19/24 \\
avg. success rate 
& 0.04 & 0.19  & 0.25 & \textbf{0.35}
& 0.40 & \textbf{0.42}
& 0.18 & 0.29  & \textbf{0.53} \\
\bottomrule
\end{tabular}
}
\vspace{4mm}
\end{table}


\section{Related Work}

\subsection{Protein Language Models}
There is growing interest in developing protein LMs at the scale of evolution, such as the series of ESM~\citep{rives2019esm,lin2022esmfold}, TAPE~\citep{rao2019evaluating}, 
ProtTrans~\citep{elnaggar2021prottrans}, PRoBERTa~\citep{nambiar2020transforming}, PMLM~\citep{he2021pre}, 
ProteinLM~\citep{xiao2021modeling}, 
PLUS~\citep{min2021pre}, 
Adversarial Masked LMs~\citep{mcdermott2021adversarial}, 
ProteinBERT~\citep{brandes2022proteinbert}, 
CARP~\citep{yang2022convolutions} in masked language modeling (MLM) paradigm, 
ProtGPT2~\citep{ferruz2022protgpt2} in causal language modeling paradigm, and several others~\citep{melnyk2022reprogramming,madani2021deep,unsal2022learning, nourani2021tripletprot, lu2020self, sturmfels2020profile, strodthoff2020udsmprot}.
These protein language models exhibit remarkable generalization ability on various downstream tasks and be able to capture evolutionary information about secondary and tertiary structures from sequences alone.
Meanwhile, recent study shows these models' potency in revealing protein structures~\citep{lin2022esmfold}, predicting the effect of sequence variation on function~\citep{meier2021language}, antibody infilling~\citep{melnyk2022reprogramming} and many other general purposes~\citep{rives2019esm}.
Simultaneously, \citet{verkuil2022language} demonstrate that the large scale protein LMs can generate \textit{de novo} proteins by generalizing beyond natural proteins, both theoretically and experimentally validating their hypothesis in exhaustive detail, in which protein LMs demonstrate competency in designing protein structure despite being exclusively trained on sequences.

\subsection{Protein Structure Generative Models}
Diffusion models have become popular tools in structural biology for protein generation, and their utility has been demonstrated across a range of generative tasks in recent years. \citet{trippe2022diffusion}, along with others, have introduced several diffusion model variants, each with its unique approach. For instance, while some models focus on generating the protein backbone by diffusing over protein coordinates, others, such as those proposed by \citet{wu2022high}, target inter-residue angles. \citet{lin2023generating} and \citet{yim2023framediff} have developed models that handle both the position and orientation of residue frames.
RFDiffusion~\citep{watson2023RFdiffusion} is a model that assists in designing protein structures for specific functions, such as enzymes. It is versatile in protein design and has been used to create therapeutic proteins, with some designs being confirmed in the laboratory.
ProteinSGM~\citep{lee2022proteinsgm} is a model that uses 2D matrices, which represent the distances and angles between protein parts, to create 3D protein structures for novel protein designs.
FoldingDiff~\citep{wu2022protein} is a model that generates protein sequences expected to fold into a specific structure. These sequences are verified with prediction tools, although they have not been experimentally confirmed yet.
Chroma~\citep{ingraham2023chroma} is a model designed for creating large proteins and protein complexes, considering various constraints like distances and symmetry. It transforms a collapsed polymer into protein backbone and sequence more quickly than older methods, thereby allowing for the efficient generation of large structures.
Multiflow~\citep{campbell2024generative} develop mulitmodal flow matching for protein structure-sequence co-generation~\citep{jin2021iterative,shi2022protein}.
ProtPardelle~\citep{chu2024all} propose an all-atom generative approach for co-design.

\end{document}